# CapsF: Capsule Fusion for Extracting psychiatric stressors for suicide from twitter

Mohammad Ali Dadgostarnia, Ramin Mousa, Saba Hesaraki

*Abstract*—Along with factors such as cancer, blood pressure, street accidents and stroke, suicide has been one of Iran's main causes of death. One of the main reasons for suicide is psychological stressors. Identifying psychological stressors in an at-risk population can help in the early prevention of suicidal and suicidal behaviours. In recent years, the widespread popularity and flow of real-time information sharing of social media have allowed for potential early intervention in large-scale and even small-scale populations. However, some automated approaches to extract psychiatric stressors from Twitter have been presented, but most of this research has been for non-Persian languages. This study aims to investigate the techniques of detecting psychi- atric stress related to suicide from Persian tweets using learning- based methods. The proposed capsule-based approach achieved a binary classification accuracy of 0.83.

*Index Terms*—Psychiatric Stressors, Persian text analysis, Capsule Network, Hybrid model, Fusion model

## 0.1. INTRODUCTION

SUICIDE is one of the leading causes of death in Iran [1]. The new statistics of the annual death registry show that in the first half of 2022, more than 290,000 deaths occurred in Iran. This is even though in the same period of the years before the certificate, the death rate fluctuated around 181 thousand people in the first six months of the year [2]. The archive of annual mortality statistics shows that the number of deaths in the first half of 2017 was about 18,000; from 2018 to 2020, it was 181,000, 182,000, and 186,000, respectively. Statistics from the Ministry of Health indicate that in the first six months of 2022, more than 56,000 people and in the first half of 2021, about 23,000 people have died. Some officials, including Iran's Medical System Organization, have stated that the death toll is two to three times the official figure. Statistics from the Ministry of Health show that 118,000 people have died from this virus in the country since the introduction of the currency. On the other hand, the statistics of the civil registry indicate that a total of 507,000 deaths were recorded in the country last year, about 130,000 more than the average number of deaths from 2018 to 2020 [2].

In the prevention and control of suicide, identifying risk factors, stressors, and causes of suicide is a fundamental step [3]. The reasons and behaviours of suicide can be divided into stressful and motivating factors. Psychological stressors, psychosocial, economic or environmental factors are the main causes of suicide that can deeply affect people's cognition, emotions and behaviour [4]. The causes of suicide and suicidal behaviours can be complex and vary significantly from one person to another, from one society to another, and from one country to another. Identifying psychiatric stressors is very important to understand the causes of possible suicidal behaviours for a particular individual, which is necessary to provide an appropriate and accurate intervention strategy. For example, in Iran, "Mehrab Ghasad Sada" as a rapper has significantly impacted the causes of suicide among young people. At the same time, this singer has no role in the suicide of Iran's neighbouring countries (Iraq, Turkey, Azerbaijan, etc). Mentions of stressors are often included in narratives such as clinical text or social media posts, especially Twitter, and therefore should be identified first for further research. Usually, people who commit suicide show different behaviours in these networks, which are usually spread in the form of conceptual posts with specific hashtags. Manual checking of this amount of information is not possible. Fortunately, advances in machine learning and natural language processing (NLP) have provided great opportunities to access mental health issues from large-scale narrative data. The flow of real-time information sharing on social media enables early detection and potential intervention for at-risk users. Most previous studies focused on analyzing the relationship between suicidal ideation and content and linguistic features such as lexical analysis [5] [6] [7] [8] [9] on social media platforms. (Twitter and Facebook). Some recent efforts have attempted to analyze tweets based on levels of distress [3], concerns [10] [11] or types of suicidal communication [12] using machine learning or classification-based approaches. However, studies have yet to be conducted to extract risk factors from social media. Authors tracked suicide risk factors from Twitter using keyword-based methods [13].

As an extension of the article, we limited the focus of this article to the automatic identification of psychiatric stress factors related to suicide and the classification of tweets related to these factors in Iran. We proposed a complete deep learning-based pipeline that uses capsule networks to extract psychiatric stressors for suicide from Twitter data. This pipeline first collects and filters tweets related to suicide using keywords, then based on a proprietary procedure, each of these tweets is given a positive and negative highlight so that the primary models can be trained.

Then, we propose a classifier based on a capsule neural network (CapsuleNet) for this purpose. The proposed capsule network uses an IndRNN model to extract initial vectors. The basis of the proposed approach is given below.
words related to suicide by GetOldTweets.

2. Emotion vector extraction by TextBlob.
3. Mapping tweets to the embedded fields.
4. Capsule Fusion training.

## 0.2. RELATED WORKS

Several research types have been conducted to identify psychiatric disorders and suicidal factors in social media data. Our discussion focuses on the literature related to data collection, feature extraction, and diagnosis of mental disorders and mortality factors. In the following, the related literature has been reviewed.

The authors in [14] discussed classifying chronic and mental diseases using social media posts. Traditional methods of identifying mental illnesses require sufficient historical data or regular monitoring of patient activities to identify patients associated with mental illness. In order to address this issue, they proposed a method to classify patients with chronic mental illnesses (such as anxiety, depression, bipolar, and ADHD) based on data extracted from Reddit. The proposed method uses a training technique. Joint (a type of semi-supervised learning approach) is used by combining the discriminating power of widely used classifiers such as random forest (RF), support vector machine (SVM) and naive. This study uses the inverse term frequency document frequency feature (TF-IDF), and technique classification algorithms used Support Vector Machine (SVM), Random Forest (RF), Logistic Regression (LR) and Simple Bayes (NB). This study uses only linguistic features to identify chronic mental illnesses, and with the co-learning approach for diagnosis of anxiety, depression and bipolar disorders using SVM obtained f-Score equal to 84, 67 and 70%, respectively.

Another similar study was conducted in [15] . The authors proposed a model to classify the severity level of psychiatric symptoms (such as depression, anxiety, and bipolar) based on data extracted from Twitter. The model was trained by integrating the linguistic features of Document Frequency Inverse Frequency (TFIDF) with N-gram weights (unigram, Bigram, and Trigram) and word2vce, with the life pattern feature (PLF) that polarity, subjectivity, and gender. The experiment was conducted by combining these features and machine learning classifiers: Support Vector Machine (SVM), Logistic Regression (LR), Random Forest (RF) and Simplest Bayes (NB). Experimental results show that SVM with Unigram-weighted TFIDF features combined with PLF performed best among other tested approaches with an accuracy score of 97.3%. Also, the SVM approach reached 96% accuracy using Bigram and 89.5% accuracy using Trigram.

The utility of artificial intelligence in suicide risk prediction and the management of suicidal behaviours was investigated in [16]. This study studied two categories of machine learning algorithms, including supervised and unsupervised approaches to identify and predict suicide. They also examined examples of conversational agents currently available for managing moodiness and suicidal behaviours, such as ElIZA[1], Ellie[2], Joy[3], Karima[4], Koko Bot[5], SPARX-R[6], Tess AI[7], Woebot[8], and Wysa[9]. They explained that at the individual level, predictive analytics by AI would help identify people in crisis to intervene with emotional support, crisis and psychoeducational resources, and alerts for emergency help. At the population level, algorithms can identify at-risk groups or suicide hotspots, helping to mobilize resources, modify policies, and support efforts. Artificial intelligence can also be suitable for supporting the clinical management of suicide in diagnosis and assessment, medication management and providing behavioural therapy. Artificial intelligence in suicide care can have several advantages, including being a time- and resource-effective alternative to clinically based strategies, adaptable to different settings and demographics, and suitable for use in remote locations with limited access to mental health supports.

In [17], the authors investigated the effect of the word processing method on mood classification. Performance is summarized in two types of classification and measurement activities. It considers the classification performance of pre-processing methods using different features and classifiers on the Twitter dataset. The pre-processing used to remove URLs and meaningless numbers or words was investigated as an essential step in this research. Therefore, Twitter data is mined, and the state for tweets on a particular topic is calculated. This research focuses on tweets about mental health problems caused by social media platforms. They calculated and analyzed attitudes from tweets using machine learning algorithms and implemented machine learning algorithms, including Naive Bayes, Random Forest, Regression, and Support Vector Machine. Function extraction methods used in the combination of machine learning could achieve a maximum accuracy of 0.92.

A machine learning approach that predicts future risk for suicidal ideation from social media data was introduced in [18]. Their main goal was to produce an algorithm called "Sui- cide Artificial Intelligence Prediction Heuristic (SAIPH)" that can predict the future risk of suicidal thoughts by analyzing public Twitter data. They trained a set of neural networks on Twitter data tested against psychological constructs associated with suicide, including burden, stress, loneliness, hopelessness, insomnia, depression, and anxiety. Using 512,526 tweets from $N$ = 283 suicidal ideation (SI) cases and 3,518,494 tweets from 2655 controls, they trained a neural network model and then ran a random forest model on the neural network outputs to predict SI status. They taught binary. The model predicted $N$ = 830 SI events from an independent pool of 277 suicidal ideators relative to $N$ = 3159 control events in all non-SI subjects with an AUC of 0.88. They validated their model using regionally obtained Twitter data and found a significant association of SI algorithm scores with county-level suicide mortality rates over 16 days in August and October 2019.

---

[1] http://www.masswerk.at/
[2] http://ict.usc.edu/prototypes/simsensei/
[3] www.hellojoy.ai/
[4] https://x2.ai/
[5] www.kik.com/casestudy/koko/
[6] www.sparx.org.nz/
[7] https://x2.ai/
[8] https://woebot.io/
[9] www.wysa.io

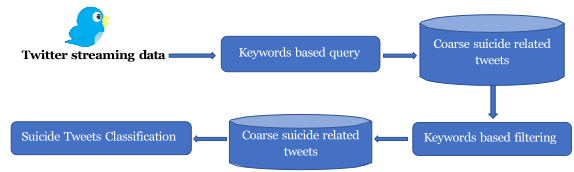

Figure 0.1: pipeline for detecting psychiatric stressors from Twitter.

Moreover, they concluded that suicide was more common in younger people. According to the authors, algorithmic approaches such as SAIPH have the potential to identify an individual's future SI risk. They can be easily adapted as clinical decision-making tools to aid in suicide screening and risk monitoring using existing technologies.

## 0.3. SYSTEM OVERVIEW

As shown in Figure 0.1, the pipeline for identifying psychiatric stressors from Twitter consists of several steps. First, we retrieve a set of suicide-related tweets using suicide-related keywords. Here, we define suicide-related tweets collected by a series of keywords as tweets that contain possible thoughts of suicide, history or plan of suicide, etc., for Twitter users. Second, we created a set of suicide-related tweets (the golden tweet sets and the models are trained on them) by filtering tweets with prominent stop words in a manual, rule-based step. Third, considering that the collection of tweets based on keywords generates a lot of noise, we use a deep learning-based classification model to select suicide-related tweets further. In particular, the approach based on capsule fusion is applied at this stage. Finally, we mentioned suicide-related tweets (golden tweets set) created from the previous step as psychological stressors.

### A. Collecting and filtering tweets

We manually compiled a list of keywords and phrases related to suicide (stressors) to collect public tweets in Farsi related to suicide from April 10, 2017, to June 26, 2022, through the Twitter API. For this purpose, we used two famous APIs GetOldTweets[10] and Tweepy[11]. A list of keywords and phrases containing 90 keywords/phrases, such as "suicide", "suicide", "I want to die", etc. was considered for this purpose. While manually reviewing the collected tweets, we found that most were about news or promotions rather than personal conversations, leading to misclassification of the data. Idea or Experience If we use the collected tweets directly, it creates more annotation load and reduces the performance of the machine learning system. To create a modified candidate set of more accurate suicide-related tweets, we removed all tweets that contained URLs or keywords such as "suicide bomb," "suicide attack," etc., during the tweet collection period. We inductively generated a list of stop words to filter tweets. By doing this, we saw a huge increase in relevant tweets, which is useful for deep learning and evaluation. The complete list of learning systems and stop keywords is given in Table 0.3.1. After filtering, 1,614,548 tweets were collected during this time period.

### B. Annotating tweets

Information was annotated to identify suicide-related tweets from the tweets. We annotated actual suicide tweets with both positive and negative tags. Positive means that the tweet is about the Twitter user's suicide or suicidal thoughts (personal experience or feeling); For this purpose, TextBlob[12] played the builder role. Tweets with a negative tag can be considered under the following conditions:

1. Lack of connection with suicide or suicidal thoughts.
2. Denial of suicide or suicidal thoughts (for example, I will not commit suicide).
3. Discussing suicide or other people's suicidal thoughts.

For this purpose, manual rules were used.

### C. Capsule Fusion based binary classification to recognize suicide related tweets

Most neural networks transform input vectors into output vectors by combining matrix multiplications and nonlinear functions, where the nonlinearities are almost always pointwise; the nonlinearities perform the same operation independently on each vector element. Usually, intermediate representation elements are referred to as the activation of a neuron that follows the neural pattern in the brain. Capsule networks have fundamental differences from standard neural networks, both recurrent and non-recurrent. In capsule networks, a set of predefined neurons in a layer is called a capsule, and usually, a capsule is sent to the next layer instead of a scalar entity in each layer. The combined activation of neurons in a capsule is called a state. Unlike traditional networks, the capsule network

---

[10]https://pypi.org/project/GetOldTweets3/
[11]https://pypi.org/project/tweepy/
[12]https://pypi.org/project/textblob/

Table 0.1: Statistics of extracted tweets based on complete keywords and phrases.

| After preprocessing | Before preprocessing | Keywords | After preprocessing | Before preprocessing | Keywords |
|---|---|---|---|---|---|
| 5001 | 5223 | ¢alF (weapon) | 1578 | 1888 | Ø,a ©9E, (death wish) |
| 24 | 24 | £dnw/©Aj9, (Lethal drugs) | 67190 | 74981 | {Œ (spite) |
| 9001 | 10061 | gntJ (Homesick) | 450 | 490 | ¢twkwJ (heartbroken) |
| 59019 | 61259 | ¢l,s (crying) | 102 | 112 | 9 HI (Dis Love) |
| 56879 | 64321 | zi9 ˉI (hanging) | 58129 | 61260 | Bz, Q,ė (Celphos) |
| 298 | 324 | ˘w¤nī wi (respiratory arrest) | 120 | 129 | ©dm˘˘blė wi (Intentional cardiac arrest) |
| 308 | 356 | Ø,a W (young death) | 380 | 398 | pA/Az W (failed young man) |
| 501 | 547 | vAnb A g¤ (Choking with a rope) | 507 | 569 | EA$ A g¤ (Gas suffocation) |
| 87998 | 10090 | W/ W (Suicide) | 71029 | 78901 | Zt W (self-harm) |
| 83732 | 87029 | ©EwF W (self-immolation) | 452 | 476 | 9,ta , W/ W (Suicide in the subway) |
| 511 | 567 | Mwa Ø,a (rat poison) | 55343 | 56718 | £dnw/ (deadly) |
| 5109 | 5781 | F9E, Ø,a (Death is precious) | 4982 | 5782 | dm Ø,a (intentional death) |
| 20123 | 60990 | , Ø,a (down with) | 6010 | 6791 | ,9 Ø,a (fatal) |
| 6531 | 7891 | FAw Ø,a (Emotional death) | 4011 | 4213 | , 9d Ø,a (Death without pain) |
| 56422 | 78911 | E Ø, (stroke) | 76590 | 10092 | pA/Az (failed) |
| 81442 | 89032 | ,a ©Aj £, (ways to die) | 16533 | 17929 | W/ W ©Aj £, (Ways to commit suicide) |
| 60192 | 81290 | £dnw/ œF (deadly poison) | 2098 | 3211 | ,a M9, (way to die) |
| 29081 | 32421 | a ¢I (end of the line) | 27891 | 32516 | ,wzAyF (cyanide) |
| 40091 | 48921 | E • yI (to shave) | 46781 | 54423 | , vAnb (Corded) |
| 76588 | 100091 | ,jE (venom) | 422 | 478 | Ø, E (stroke) |

"bomb", "suicide attack", "suicide attacks", "Bus attack", "Bus attacks", "car attack", "car attacks", "suicide hotline", "https://", "http://", ".ac.ir"    Stop Keywords sends features as a feature vector(a capsule) to the next layer in each layering; in typical networks, this feature is a scalar value.

[19] defined a capsule as a group of neurons with sampling parameters represented by activity vectors, where the length of the vector represents the probability of feature presence. Capsule networks consist of the convolutional/RNN, primary capsule (PC) and class capsule layers. The capsule network structure is such that the initial capsule layer is the first layer of this network, followed by an unknown number of capsule layers and activator functions until the last capsule layer. The convolution layer extracts the feature from the image, and the output is entered into the initial capsule layer. Each capsule $i$ (where $1 \leq i \leq N$) in layer $l$ has an activity vector $u_i \in R$ to encode spatial information in the form of sample parameters. The output vector $u_i$ of the lower-level capsule $i$ is fed to all capsules in the next layer $l+1$. The $j-th$ capsule in layer $l+1$ receives $u_i$ and finds its product with the corresponding weight matrix $W_{ij}$. The resulting vector $\hat{u}_{j|i}$ is capsule $i$ in the transformation of level $l$ of the entity represented by capsule $j$ at level $l+1$. The prediction vector of a PC, $\hat{u}_{j|i}$, shows how much the initial capsule $i$ contributes to the class capsule $j$. In fact, the vector $\hat{u}_{j|i}$ is obtained through the following relation [19]:

$$\hat{u}_{j|i} = W_{ij}u_i \quad (1)$$

Where $u_i$ is the initial vector taken from the convolution layer (in some models, this vector is also taken from RNN networks) and $W_{ij}$ is the random weight matrix. The product of the prediction vector and the coupling coefficient agreeing on these capsules is performed to obtain the prediction of an initial capsule I to the capsule of class $j$. If the agreement is high, then the two capsules are related. As a result, the coupling coefficient increases and otherwise decreases. The weighted sum ($S_j$) of all these initial capsule predictions for capsule class $j$ is obtained according to the following equation.

$$S_j = \Sigma_{i=1}^N C_{ij}\hat{u}_{j|i} \quad (2)$$

Figure 0.2 shows a general representation of the process of a capsule.

Next, the squashing function is applied to $S_j$.

$$v_j = \frac{||S_j||^2 S_j}{1+||S_j||^2/||S_j||_j} \quad (3)$$

The squashing function is theoretically unmotivated, but intuitively, it further reduces the norm of parent capsules that have a smaller norm. One way to see the squashing function is that it increases the convergence speed of the routing algorithm. The squashing function ensures that the output length from the capsule is like a probability between 0 and 1. $v_j$ is routed from one capsule layer to the capsule of the next layer and treated in the same way as discussed. The coupling coefficient $c_{ij}$ ensures that prediction $i$ at level $l$ is related to $j$ at layer $l+1$. During each iteration, $c_{ij}$ is sub-updated by finding the dot product of $\hat{u}_{j|i}$ and $v_j$ [19].

$$c_{ij} = \frac{exp(b_{ij})}{\Sigma_k exp(b_{ik})}$$

In particular, the vector values associated with each capsule can be viewed as a part of two numbers. A probability that represents the existence of a feature that a capsule encapsulates and a set of instantiation parameters that can serve as an explanation of compatibility between layers. Thus the path associated with agreement derives from the fact that when lower-level capsules agree on a higher-level layer capsule, they establish a "one-part-of-a-whole" relationship that represents a path relationship. This algorithm is called dynamic routing with the agreement. Since low-level capsules represent the core entities of an object, we need a way to pass information to the appropriate parent capsule that represents the correct entity at the next layer. A dynamic routing algorithm, consensus routing, makes this possible, which decides the outputs of lower-level capsules that should go to the next layer through an iterative process. This mechanism is based on a prediction by the lower-level capsule for the deployment parameter of the higher-level capsule. This prediction is calculated through a transformation matrix and is activated when multiple predictions from low-level capsules match the output of the high-level capsule. The length of the output vector gives the probability of the entity represented by the capsule. A non-linear squashing function ensures that the length stays in the range of 0 to 1, reducing short vectors to approximately zero and long vectors to approximately one [19].

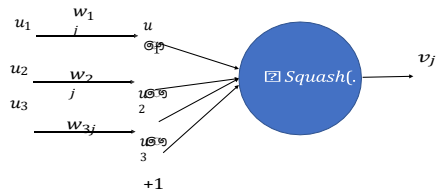

Figure 0.2: Overview of the process of a capsule.

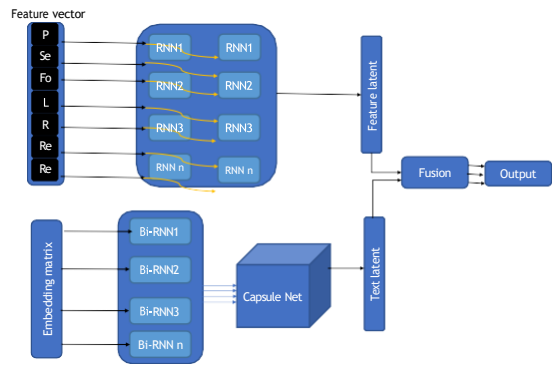

Figure 0.3: Capsule Fusion model overview.

## 0.4. METHODOLOGY

The general goal of the proposed approach is to use information-based and extracted emotional features along with tweet features. In general, non-based features such as the sentiment of the tweet, the number of followers of the person who sent the tweet, etc. can help as auxiliary information in more correct classification. An overview of the proposed approach is shown in Figure 0.3. This approach has two inputs. In the following, we will discuss each component in more detail.

**Feature vector:** This vector is a vector of extracted emotional features and tweets that shows the influence of each person separately. This vector is given as input to an RNN network. This vector contains the features that Get old Tweet extracts for each ID. These features include:

1. **Sentiment(Se):** takes the value of 0, -1 and 1 for neutral, negative and positive, respectively.
2. **Polarity (p):** Polarity is between [-1,1]; -1 defines a negative feeling and 1 a positive feeling.
3. **Subjectivity(s):** subjectivity quantifies the number of personal opinions and factual information in the text. Higher subjectivity means that the text contains personal opinions rather than factual information.
4. **Follower count (Fo):** specifies the number of followers of the author of the tweet.
5. **Likes(L):** Specifies the number of likes for the tweet.
6. **Replies(R):** Specifies the number of replies for that tweet.
7. **Retweets (Re):** Specifies the number of retweets of that tweet.

These features are given to the network under the name Feature vector. An overview of these features is given in Figure 0.4. The output of this part is considered a latent feature.

**CapsuleNet:**

The hidden textual features in the model are extracted using a Bi-GRUCapsule network. CapsuleNets are mainly used in computer vision tasks, such as image classification [20] or object detection [21]. CapsuleNet has also achieved significant performance in many natural language processing (NLP) tasks [20] [21] [22] [23] [24] [25] . With the capsule approach, the neural network can generate local features around each word from the adjacent word and combine them using a max operation to create a word-level embedding of fixed size, as shown in Figure 5.

Therefore, we use CapsuleNet to model textual hidden features for positive or negative detection. Let word $j$ in tweet $i$ be represented as $x_{i,j} \in R^k$, which is a k-dimensional embedding vector. Suppose the maximum length of the tweet is $n$. The news of less than $n$ words can be shown as a sequence of length $n$ because one of the limitations of neural networks is the constant length of inputs, and all inputs must be the same size. Hence, the overall input can be written as follows:

$$X_{i,1:n}^{Tl} = x_{i,1} \bigoplus x_{i,1} \bigoplus x_{i,2} \bigoplus ... \bigoplus x_{i,n} \quad (5)$$

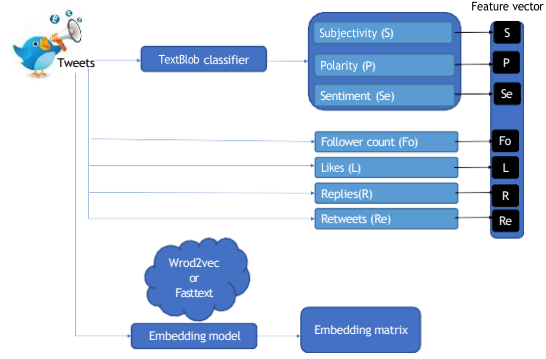

Figure 0.4: Feature vectors extracted by Tweepy, getoldtweets and TextBlob.

The desired Bi-IndRNNCapsule network includes the following layers:

1. **Bi-IndRNN layer**: The input of the Bi-IndRNN layer is the embedded vectors taken from the embedding layer. If we display these vectors as $Out_{embed} = [X_1, X_2, ..., X_n]$. Bi-IndRNN's vector at step $T$ is $X^{200}$. Then $h_t = h_1, h_2, ..., h_t$ represent the hidden states extracted from the IndRNN network, obtained through the following relationship:

$$h_t = (W_x + u^k h_{t-1} + b) \quad (6)$$

In some tasks, input sequence inversion can improve network performance. Bidirectional IndRNN networks (Bi-IndRNN) process data in both Forward and Backward directions. If $h_t$ is the Forward output for the sequence $x_1^t$, which is $t = 1, 2, 3, ..., t$ and if $\overleftarrow{h_t}$ is the Backward output for $x_1^t$ where $t = t, ...3, 2, 1$. Then the Bi-IndRNN output is obtained through the step-by-step combination of Forward and Backward outputs as $h_t = \overrightarrow{h_t}.\overleftarrow{h_t}$. This network has twice as many free parameters compared to the unidirectional mode.

2. **Capsule layer**: The features encoded by the Bi-IndRNN layer are fed into a capsule network. If the output of Bi-IndRNN is equal to $h_i$ and $W$ is the weight matrix, then $\hat{v_{ij}}$, which represents the prediction vector, is obtained from the following equation:

$$\hat{v_{ij}} = w_{ij}.h_i \quad (7)$$

The set of inputs to a capsule $s_j$ is a weighted set of all prediction vectors $\hat{v}_{ij}$ which is calculated through the following equation:

$$s_j = \sum c_{i,j}.\hat{v}_{ij} \quad (8)$$

2. **Classification layer**: At first, the output of the capsule, which we call latent text, is flattened, combined with the 1 neuron. In general, this process can be formulated as follows, where F represents the number of neurons in the last layer:

$$p = W_{Textlatent} + W_{Featurelatent} \boxtimes F \quad (9)$$

The output $P$ should be such that it represents the probability of each of the two classes. For this purpose, using Sigmoid, for each $f_i \boxtimes F$, the value of $P_i$ is calculated as follows:

$$P_i = \frac{1}{1+e^{-(f_i)}} \quad (10)$$

To determine whether it was positive or negative, it is enough to calculate Round($p_i$). If its value is 1, it is positive and if it is 0, it is negative.

0.5. EXPERIMENTAL RESULTS

The proposed method was trained on a computer with core i7-7700k CPU (4.2 GHZ), 32 GB DD RAM and NVIDIA GeForce GTX 1080 Ti. Also, the software configurations of the proposed approach are given in the table0.2:

Table 0.2: Software configuration

| Application | version |
|---|---|
| Operation system | Linux Ubu |
| Python | Version 3.6 |
| TensorFlow | Version 2.0 |
| Keras | Version 2.0 |

$$Accuracy = \frac{TP+TN}{(TP+FN+TN+FP)} \quad (11)$$

$$Precision = \frac{TP}{(TP+FP)} \quad (12)$$

$$Recall = \frac{TP}{(TP+FN)} \quad (13)$$

$$F1-score = \frac{2*precision*recall}{(precision+recall)} \quad (14)$$

where *TP* (True Positive) is the number of correct samples that have been correctly detected, *FP* (False Positive) is the number of false samples that have been correctly detected, *TN* (True Negative) is the number of correct samples that have been detected as incorrect, and *FN* (False Negative) is the number of false samples that are wrongly recognized.

## 0.4. EXPERIMENTAL RESULTS

To evaluate the proposed approach, 80% of the data were considered training data and 20% test data. Also, several different approaches were considered comparative approaches for the comprehensiveness of the research. The results of different approaches are given in Table 0.6. These approaches are divided into four approaches: Linear (numerical mapping of words and their frequencies), CNN (convolutional network as the core and basis of learning), RNN (recurrent networks as the core and basis of learning) and capsule-based approaches are categorized.

The Bag of Words approach ( [26]) reached 0.66 per- cent accuracy on the dataset. This approach reached Pre- cision=0.65, Recall=0.66, and F1=0.65 on the Positive and negative classes; this approach reached Precision=0.75, Re- call=0.78, and F1=0.76. Another approach used is the n- grams approach( [26]), which obtained an accuracy of 0.60 on the target data set. This approach obtained 0.64, 0.67, and 0.65 in the Positive class and 0.68, 0.68, and 0.68 in the Negative class, respectively, for the evaluation criteria of Precision, Recall, and F1, and obtained a weaker performance than Bag of Words. The combination of n-gram and TFIDF in n-gram TFIDF obtained better performance than n-gram. This approach reached an accuracy of 0.64 on experimental data. In the Positive class, this approach reached precision=0.67, Recall=0.68, and F1=0.67, and in the Negative class, it reached Precision=0.71, Recall=0.73, and F1=0.72.

In methods based on convolutional networks, it was tried to use the convolutional network as the main core of the model. In Char-level CNN small ( [26]), the model reached an accuracy of 0.67, higher than all linear models. This approach also reached Precision=0.68, Recall=0.69, and F1=0.68 in the Positive class and Precision=0.73, Recall=0.76, and F1=0.74 in the Negative class. Char-level CNN large ( [26]) performed much worse than the linear and Char-level CNN small ap- proaches and achieved an accuracy of 0.57. These poor results are based on the lack of accurate identification of the Positive class.

Two approaches VDCNN-29 layers ( [27]) and Word-level CNN ( [28]), obtained equal accuracy. These two approaches reached an accuracy of 0.65 due to the same inputs and data structure. VDCNN-29 layers approach ( [27]) reached Precision=0.69, Recall=0.72, and F1=0.70 in the Positive class and Precision=0.73, Recall=0.76, and F1=0.74 in the Negative class. Word-level CNN approach ( [28]) obtained values of 0.68, 0.68, and 0.68 for the Positive class and 0.72, 0.74, and 0.73 for the Negative class, respectively, for Precision, recall, and F1. The Fasttext approach ( [29]) reached a precision of 0.64 on the test data. This approach reached Precision=0.70, Recall=0.71, and F1=0.69 in the Positive class and Precision=0.73, Recall=0.73 in the Negative class. , and F1=0.73 was reached. According to different approaches based on RNN networks, only one approach was evaluated on the desired data set.

The D-LSTM approach ( [30]) reached an accuracy of 0.64 on the desired data set. This approach achieved Preci- sion=0.68, Recall=0.70, and F1=0.69 in the Positive class and Precision, Recall, and F1=0.72 in the Negative class. Capsul's approaches obtained stunning results regarding the structure of Persian language sentences. The Bi-GRUCapsule approach ( [31]) reached a precision of 0.76 on the target data that only included textual information. This approach compared with the Positive class to Precision=0.80, recall=0.85, and f1= reached
0.82 and also compared to the Negative class, this approach reached Precision=0.83, Recall=0.85, and F1=0.84. The pro- posed CapsuleFusion approach recorded the highest accuracy compared to other approaches. This approach achieved an accuracy of 0.83 on the test data, which has improved the accu-racy by 0.07 compared to the Bi-GRUCapsule approach. This approach reached Precision=0.89, Recall=0.92, and F1=0.90 in the Positive class, and also compared to the Negative class, this approach reached Precision=0.91, Recall=0.93, and F1=0.92. The proposed approach achieved a 0.19 improvement in accuracy compared to the D-LSTM approach ( [30]), 0.16 improvement compared to the best CNN-based approach, and
0.17 improvement compared to the best Linear approach.

Allocation of parameters to the model requires precise engineering of the parameters because choosing the wrong ones will cause the model not to converge. In the following, the effect of two parameters, batch size and dropout, will be investigated. Dropout is used to reduce pre-fitting in deep networks. In simpler terms, dropout refers to ignoring neurons during the training phase of a particular set of neurons that are randomly selected. By "ignore" that means these units are not considered forward or backward during a particular pass. More technically, nodes with probability 1-p are removed from the network at each training step or kept with probability p, leaving a reduced network. The input and output edges to a deleted node are also deleted. This research considered the values of 0.2, 0.3, 0.4, 0.5, 0.6, 0.7, 0.8 and 0.9 for this purpose. According to the figure on the left in Figure 0.6, the proposed approach has reached the best accuracy at 0.4. Batch size is the number of samples sent to the network simultane- ously. Choosing a large batch size leads to early coverage, but on the other hand, it also increases processing requirements. Choosing a smaller batch increases the probability of getting stuck in local minima. In the proposed approach according to the figure on the right in Figure 0.6, the batch size of 2, 4, 6, 8, 16, 32, 64, 128, 512, and 1024 was considered, and the proposed approach reached the highest accuracy in the batch size of 2 and 16.

## 0.5. CONCLUSION

Quantitative automatic methods have been proposed to extract psychiatric stressors from Twitter and Persian tweets. Most of this research was in English. Studies have been hampered mainly by the lack of annotated collections, which are time-consuming and expensive to build, and the inherent language problems created by extra-clinical stressors. This research is the first and most comprehensive attempt to extract psychiatric stressors from Persian Twitter data using deep learning and capsule network approaches. We have created an annotated Twitter collection on identifying stressors associated with suicide. We also conducted extensive experiments to justify using the presented approaches, which included convo- lutional, recursive, and capsule-based approaches. Our method has performed well in identifying tweets related to suicide and achieved an accuracy of 0.83. Due to the power of deep learning approaches, a set of features extracted from tweets can be improved to improve future works. In [32] and [33] the authors have introduced a set of user features as feature vectors that can be an improvement for future approaches.

Table 0.3: Test accuracy (%) of all the models on the datasets.

| Models | | | Precision | Recall | F1 | Accuracy |
|---|---|---|---|---|---|---|
| Linear | Bag of Words( [26]) | Positive | 0.65 | 0.66 | 0.65 | |
| | | Negative | 0.75 | 0.78 | 0.76 | 0.66 |
| | n-grams( [26]) | Positive | 0.64 | 0.67 | 0.65 | |
| | | Negative | 0.68 | 0.68 | 0.68 | 0.60 |
| | n-grams TFIDF( [26]) | Positive | 0.67 | 0.68 | 0.67 | |
| | | Negative | 0.71 | 0.73 | 0.72 | 0.64 |
| CNN | Char-level CNN small( [26]) | Positive | 0.68 | 0.69 | 0.68 | |
| | | Negative | 0.73 | 0.76 | 0.74 | 0.67 |
| | Char-level CNN large( [26]) | Positive | 0.59 | 0.63 | 0.61 | |
| | | Negative | 0.61 | 0.63 | 0.62 | 0.57 |
| | VDCNN- 29 layers( [27]) | Positive | 0.69 | 0.72 | 0.70 | |
| | | Negative | 0.73 | 0.76 | 0.74 | 0.65 |
| | Word-level CNN( [28]) | Positive | 0.68 | 0.68 | 0.68 | |
| | | Negative | 0.72 | 0.74 | 0.73 | 0.65 |
| | fastText( [29]) | Positive | 0.70 | 0.71 | 0.70 | |
| | | Negative | 0.73 | 0.73 | 0.73 | 0.64 |
| RNN | D-LSTM( [30]) | Positive | 0.68 | 0.70 | 0.69 | |
| | | Negative | 0.72 | 0.72 | 0.72 | 0.64 |
| Capsule | Bi-GRUCapsule( [31]) | Positive | 0.80 | 0.85 | 0.82 | |
| | | Negative | 0.83 | 0.85 | 0.84 | 0.76 |
| | Capsule Fusion(Capsfusion) | Positive | 0.89 | 0.92 | 0.90 | |
| | | Negative | **0.91** | **0.93** | **0.92** | **0.83** |

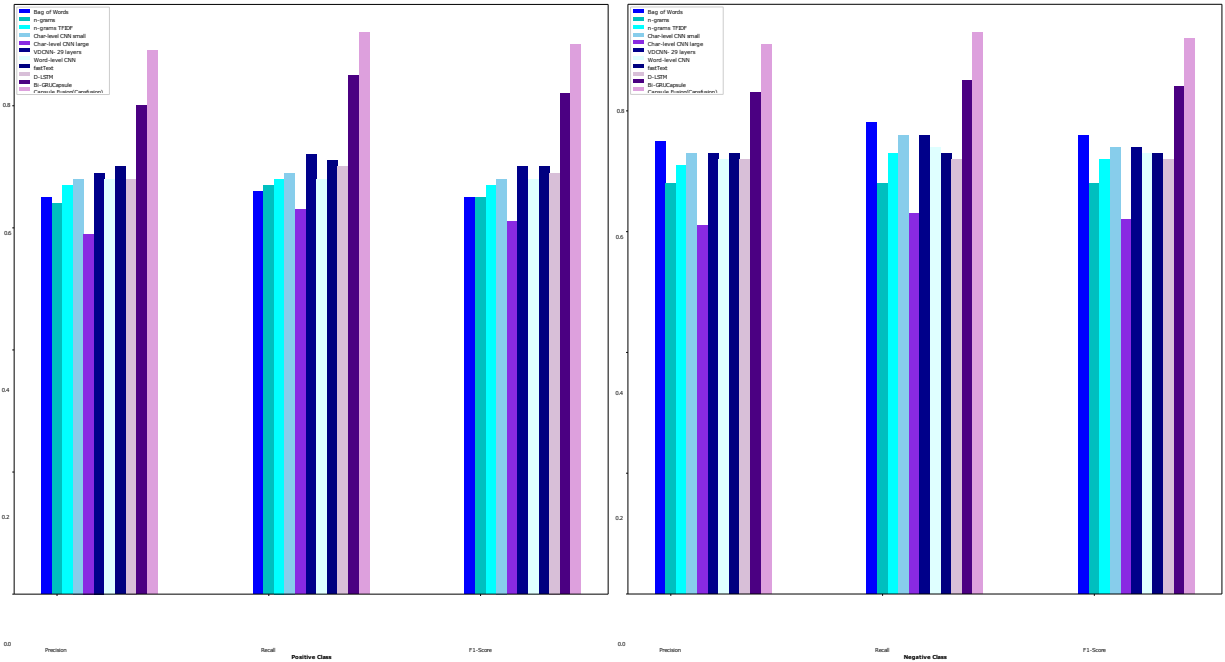

Figure 0.5: Positive and Negative Frequency.

Figure 0.6: Metrics Per Batch size and Dropout.

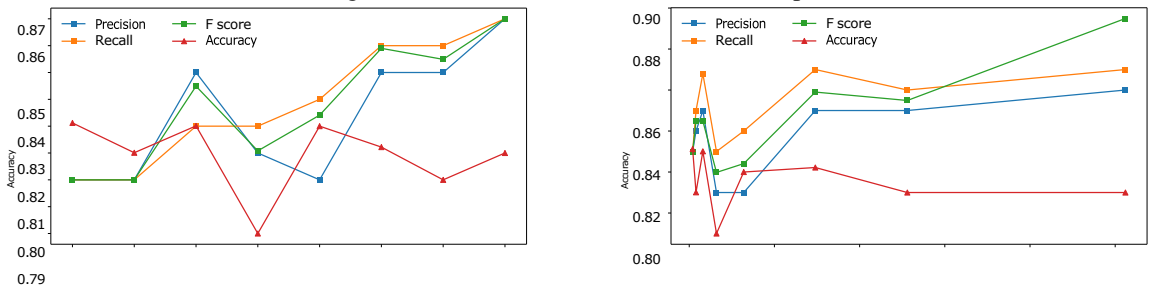

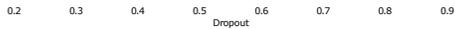